\documentclass[conference]{IEEEtran}
\IEEEoverridecommandlockouts
\usepackage{cite}
\usepackage{bm}
\usepackage{amsmath,amssymb,amsfonts}
\usepackage{graphicx}
\usepackage{textcomp}
\usepackage{xcolor}
\usepackage{booktabs}
\usepackage{caption}
\usepackage{subfigure}
\usepackage{algorithm}
\usepackage{algpseudocode}
\usepackage{multirow}
\usepackage{longtable}
\usepackage{url}

\def\BibTeX{{\rm B\kern-.05em{\sc i\kern-.025em b}\kern-.08em
    T\kern-.1667em\lower.7ex\hbox{E}\kern-.125emX}}
\begin{document}

\title{Introducing Competitive Mechanism to Differential Evolution for Numerical Optimization \\}

\author{\IEEEauthorblockN{Rui Zhong, Yang Cao, Enzhi Zhang}
\IEEEauthorblockA{\textit{Graduate School of Information Science and Technology} \\
\textit{Hokkaido University}\\
Sapporo, Japan \\
\{rui.zhong.u5, yang.cao.y4, enzhi.zhang.n6\}@elms.hokudai.ac.jp}
\and 
\IEEEauthorblockN{Masaharu Munetomo}
\IEEEauthorblockA{\textit{Information Initiative Center} \\
\textit{Hokkaido University}\\
Sapporo, Japan \\
munetomo@iic.hokudai.ac.jp}
}

\maketitle

\begin{abstract}
This paper introduces a novel competitive mechanism into differential evolution (DE), presenting an effective DE variant named competitive DE (CDE). CDE features a simple yet efficient mutation strategy: DE/winner-to-best/1. Essentially, the proposed DE/winner-to-best/1 strategy can be recognized as an intelligent integration of the existing mutation strategies of DE/rand-to-best/1 and DE/cur-to-best/1. The incorporation of DE/winner-to-best/1 and the competitive mechanism provide new avenues for advancing DE techniques. Moreover, in CDE, the scaling factor $F$ and mutation rate $Cr$ are determined by a random number generator following a normal distribution, as suggested by previous research. To investigate the performance of the proposed CDE, comprehensive numerical experiments are conducted on CEC2017 and engineering simulation optimization tasks, with CMA-ES, JADE, and other state-of-the-art optimizers and DE variants employed as competitor algorithms. The experimental results and statistical analyses highlight the promising potential of CDE as an alternative optimizer for addressing diverse optimization challenges.
\end{abstract}


\section{Introduction}  \label{sec:1}
Differential Evolution (DE) \cite{Storn:97} is a potent optimization algorithm categorized within the evolutionary algorithm (EA) family. Unlike conventional mathematical optimization methods that hinge on gradients, DE is a stochastic optimization approach inspired by the principles of natural selection and evolution \cite{Yu:23, Ren:24}. By iteratively applying crossover, mutation, and selection operations, DE refines the population of candidate solutions for a given problem, progressively steering it towards more optimal solutions \cite{Yang:22}.

Due to its simplicity, robustness, and efficiency in addressing complex optimization problems across various domains, DE is particularly adept at handling scenarios where the objective function is non-linear, non-convex, noisy, or lacks derivative information \cite{Zhong:23}. These exceptional characteristics render DE a versatile tool for addressing real-world optimization challenges encountered in engineering \cite{Wu:23, Yang:23, Zhang:24}, finance \cite{Xue:23, Zhang:22}, machine learning \cite{Zhong:23_1, Suchitra:23}, and numerous other fields \cite{Guo:22, Biswas:23, Wahab:24, Zhang:24_1}. Consequently, DE has garnered widespread attention from researchers and scholars. Concurrently, numerous variants of DE have been introduced to tackle diverse optimization tasks. While this paper does not delve into the comprehensive history and evolution of DE, readers keen on exploring this topic further can refer to \cite{Das:16, Bilal:20, Mohamad:22}.

This paper introduces a novel competitive mechanism into DE and presents a mutation operator termed DE/winner-to-best/1. By integrating this innovative mutation strategy with the basic DE optimizer, we propose a simple yet efficient variant of DE, termed Competitive DE (CDE). To thoroughly investigate the performance of our proposed CDE, we conduct a comprehensive series of fair comparison experiments on the IEEE CEC2017 benchmark functions. Furthermore, we extend CDE to address real-world engineering simulation optimization problems. Through this straightforward modification, we achieve satisfactory performance across various optimization tasks, even when competing with state-of-the-art DE variants. The experimental results and statistical analyses highlight the efficacy and versatility of CDE in tackling diverse optimization challenges.

The remainder of this paper is organized as follows: Section \ref{sec:2} introduces the framework of basic DE and engineering simulation problems. Section \ref{sec:3} introduces our proposed CDE in detail. Section \ref{sec:4} presents numerical experiments and statistical analyses, and the performance analyses are discussed in Section \ref{sec:5}. Finally, Section \ref{sec:6} concludes this paper.

\section{Basic DE}  \label{sec:2}
We begin the introduction of the basic DE by the definition of optimization problems. Without loss of generality, the minimization problem is mathematically defined by Eq. (\ref{eq:2.1}).
\begin{equation}
    \label{eq:2.1}
    \begin{aligned}
        f(\bm{x}^*) = \min f(\bm{x}), s.t. \bm{x} \in \mathcal{R}^D
    \end{aligned}
\end{equation}
where $\bm{x} = \{x_1, x_2, ..., x_D\}$ is a solution vector with $D$ dimensions. Optimization algorithms aim to find optimum $\bm{x}^*$ with a limited computational budget.

Subsequently, we outline the four primary components of DE: initialization, mutation, crossover, and selection. It's important to note that all explanations are presented within the context of the minimization.

\textbf{Initialization}: The first step of DE implementation is population initialization, which is described in Eq. (\ref{eq:2.2}).
\begin{equation}
    \label{eq:2.2}
    \begin{aligned}
        X = & \begin{bmatrix}
        \bm{x_1} \\ 
        \bm{x_2} \\
        \bm{x_3} \\
        \vdots \\
        \bm{x_N} \\
        \end{bmatrix} = \begin{bmatrix}
            x_{11} &  x_{12} & \cdots & x_{1D} \\
            x_{21} &  x_{22} & \cdots & x_{2D} \\
            x_{31} &  x_{32} & \cdots & x_{3D} \\
            \vdots &  \vdots & \ddots & \vdots \\
            x_{N1} &  x_{N2} & \cdots & x_{ND} \\
        \end{bmatrix} \\
        x_{ij} = & r \cdot (\bm{ub}_j - \bm{lb}_j) + \bm{lb}_j
    \end{aligned}
\end{equation}
where $\bm{x}_i$ denotes the $i^{th}$ individual and $x_{ij}$ represents the value in the $j^{th}$ dimension of the $\bm{x}_i$. $\bm{lb}_j$ and $\bm{ub}_j$ are the lower and the upper bound of the $j^{th}$ dimension, respectively, and $r$ is a random number in (0, 1). 

\textbf{Mutation}: When DE enters the main loop, the mutation operation is first activated to construct the mutant vector, and the representative mutation schemes are listed in Eq. (\ref{eq:2.3}).
\begin{equation}
    \label{eq:2.3}
    \begin{aligned}
        & \text{DE/rand/1}: \bm{v}^t_i = \bm{x}^t_{r1} + F \cdot (\bm{x}^t_{r2} - \bm{x}^t_{r3}) \\
        & \text{DE/cur/1}: \bm{v}^t_i = \bm{x}^t_i + F \cdot (\bm{x}^t_{r1} - \bm{x}^t_{r2}) \\
        & \text{DE/best/1}: \bm{v}^t_i = \bm{x}^t_{best} + F \cdot (\bm{x}^t_{r1} - \bm{x}^t_{r2}) 
    \end{aligned}
\end{equation}

where $\bm{x}^t_{r1}$, $\bm{x}^t_{r2}$, and $\bm{x}^t_{r3}$ are randomly selected individuals from the population and mutually distinct in the $t^{th}$ iteration, $\bm{x}^t_{best}$ denotes the best solution found so far, and $F$ is the scaling factor to control the amplification of differential vector.

\textbf{Crossover}: Although many novel crossover strategies such as exponential crossover \cite{Meng:23} and blending crossover \cite{Ghosh:17} have been proposed, the most commonly utilized binomial crossover is expressed in Eq. (\ref{eq:2.4}).
\begin{equation}
    \label{eq:2.4}
        \bm{v}^t_{i, j} = \begin{cases}
            \bm{u}^t_{i, j}, \ if \ r \leq Cr \ or \ j=j_{rand} \\
            \bm{x}^t_{i, j}, \ otherwise
        \end{cases}
    \begin{aligned}
    \end{aligned}
\end{equation}
$Cr$ represents the crossover rate to control the probability of inherited genes between the mutant vector $\bm{u}^t_{i, j}$ and the parent individual $\bm{x}^t_{i, j}$. $j_{rand}$ is a random integer in $\{1, 2, ..., D\}$. 

\textbf{Selection}: The selection mechanism in basic DE ensures the survival of elite individuals to the next iteration, as formulated in Eq. (\ref{eq:2.5}).
\begin{equation}
    \label{eq:2.5}
        \bm{x}^{t+1}_{i, j} = \begin{cases}
            \bm{u}^t_{i, j}, \ if \ f(\bm{u}^t_{i, j}) \leq f(\bm{x}^t_{i, j}) \\
            \bm{x}^t_{i, j}, \ otherwise
        \end{cases}
    \begin{aligned}
    \end{aligned}
\end{equation}

The one-to-one greedy selection mechanism in DE can survive the elites while maintaining population diversity. In summary, the pseudocode of the basic DE is presented in Algorithm \ref{alg:2.1}.
\begin{algorithm}
\caption{Basic DE}\label{alg:2.1}
\begin{algorithmic}[1]
\Require Population size:$N$, Dimension:$D$, Max. iteration:$T$
\Ensure Optimum: $\bm{x}^t_{best}$
    \State $X \gets \textbf{initial}(N, D)$ \# Population initialization 
    \State $t = 0$ 
    \State $\bm{x}^t_{best} \gets \textbf{best}(X)$ 
        \While {$t<T$} 
            \For {$i=0 \ to \ N$}
                \State Construct the mutant vector using Eq. (\ref{eq:2.3})
                \State Crossover using Eq. (\ref{eq:2.4})
            \EndFor
            \State Selection using Eq. (\ref{eq:2.5})
            \State $\bm{x}^t_{best} \gets \textbf{best}(X)$
            \State $t \gets t+1$
        \EndWhile
        \State \textbf{return} $\bm{x}^t_{best}$
\end{algorithmic}
\end{algorithm}

\section{Competitive Differential Evolution (CDE)}  \label{sec:3}
\begin{figure}[htbp]
    \centering
    \includegraphics[width=6cm]{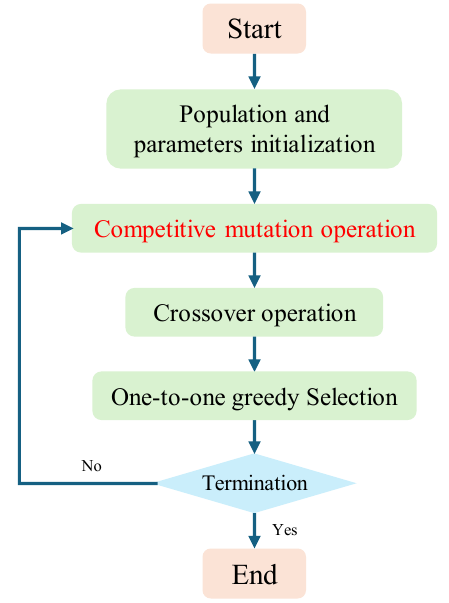}
    \caption{The flowchart of CDE.}
    \label{fig:3.1}
\end{figure}

This section introduced the proposed CDE in detail. Based on the simple yet effective architecture of DE, the main flowchart of CDE is presented in Fig. \ref{fig:3.1}. The novel component in CDE is highlighted in red. 

We introduce the competitive mechanism to CDE and propose a novel DE/winner-to-best/1 mutation operation, as formulated in Eq. (\ref{eq:3.1}).
\begin{equation}
    \label{eq:3.1}
    \begin{aligned}
        \bm{v}^g_i = \begin{cases}
            \bm{x}^g_{r1} + F_1 \cdot (\bm{x}^g_{best} - \bm{x}^g_{r1}) + F_2 \cdot (\bm{x}^g_{r2} - \bm{x}^g_{r3}), \ (a) \\
            \bm{x}^g_i + F_1 \cdot (\bm{x}^g_{best} - \bm{x}^g_{i}) + F_2 \cdot (\bm{x}^g_{r2} - \bm{x}^g_{r3}), \ (b)
        \end{cases}
    \end{aligned}
\end{equation}
where $F_1$ and $F_2$ are two random values following a normal distribution $N(0.5, 0.3)$ as recommended in \cite{Qin:05}. Simply, the proposed DE/winner-to-best/1 strategy randomly selects a competitor individual $\bm{x}^g_{r1}$ first, and if it has a better fitness value, $\bm{x}^g_{r1}$ will replace the current individual $\bm{x}^g_i$ and act as the base vector in the mutation operator to construct the mutant vector $\bm{v}^g_i$ using the DE/rand-to-best/1 scheme, as expressed in Eq. (\ref{eq:3.1}) (a); otherwise, the current individual $\bm{x}^g_i$ will survive and the DE/cur-to-best/1 mutation scheme in Eq. (\ref{eq:3.1}) (b) is activated to construct the mutated vector $\bm{v}^g_i$. 

The structure of the proposed DE/winner-to-best/1 strategy resembles a fusion of the DE/cur-to-best/1 and DE/rand-to-best/1 strategies. However, incorporating a competitive mechanism enables the automatic selection of the most suitable mutation strategy, ensuring the utilization of superior knowledge to construct the mutated vector. Furthermore, CDE can benefit from the proposed DE/winner-to-best/1 mutation strategy from (i). Strengthened convergence: By leveraging a superior base vector, the optimization process experiences rapid convergence and contributes to accelerating the approach to optimal solutions. (ii). The prevention of premature optimization: The inclusion of the random differential vector $F_2 \cdot (\bm{x}^g_{r2} - \bm{x}^g_{r3})$ serves to mitigate premature optimization and promote the exploration of diverse search spaces.  (iii). Versatile scaling factor: The utilization of a simple yet effective random generator-based scaling factor $F_i$ facilitates the generation of differential vectors with varying scales, thereby enhancing the adaptability and robustness of the mutation strategy.

Furthermore, the conventional constant crossover rate in basic DE is replaced by a random value sampled from a normal distribution $N(0.5, 0.3)$, as recommended in \cite{Jia:23}. This simple modification in CDE yields two significant benefits. First, it enhances the balance between exploration and exploitation. By introducing a randomly chosen crossover rate, CDE can explore a broader search space by occasionally performing crossover operations, potentially leading to the discovery of new promising solutions. Simultaneously, it retains the capability to exploit current best solutions by occasionally abstaining from crossover operations. Second, this modification strengthens the robustness of CDE. Introducing randomness into the crossover rate helps prevent the algorithm from becoming trapped in specific regions of the search space, thereby improving its robustness when tackling complex optimization problems.

\section{Numerical experiments}  \label{sec:4}

This section introduces the details of the designed numerical experiments to evaluate the performance of CDE. Section \ref{sec:4.1} details the experimental setting, and Section \ref{sec:4.2} presents the experimental results and statistical analyses for further discussion.

\subsection{Experiment settings} \label{sec:4.1}

\subsubsection{Benchmark functions} \label{sec:4.1.1}
We conduct comprehensive numerical experiments on CEC2017 benchmark functions and six engineering simulation optimization tasks \cite{Zhong:24_1}. The details of engineering problems are presented in the following contexts. These benchmarks are accessed via the OpFuNu library \cite{Thieu:20} and the ENOPPY library \cite{Thieu:23} using Python 3.11.

Six engineering simulation models adopted in our numerical experiments include cantilever beam design (CBD), corrugated bulkhead design (CBHD), gear train design (GTD), three-bar truss design (TBTD), tubular column design (TCD), and welded beam design (WBD). In the following mathematical models, $f(x)$ denotes the objective function, and $g_i(x)$ represents the $i^{th}$ constraint function.

\begin{figure*}[htbp]
    \centering
    \includegraphics[width=16cm]{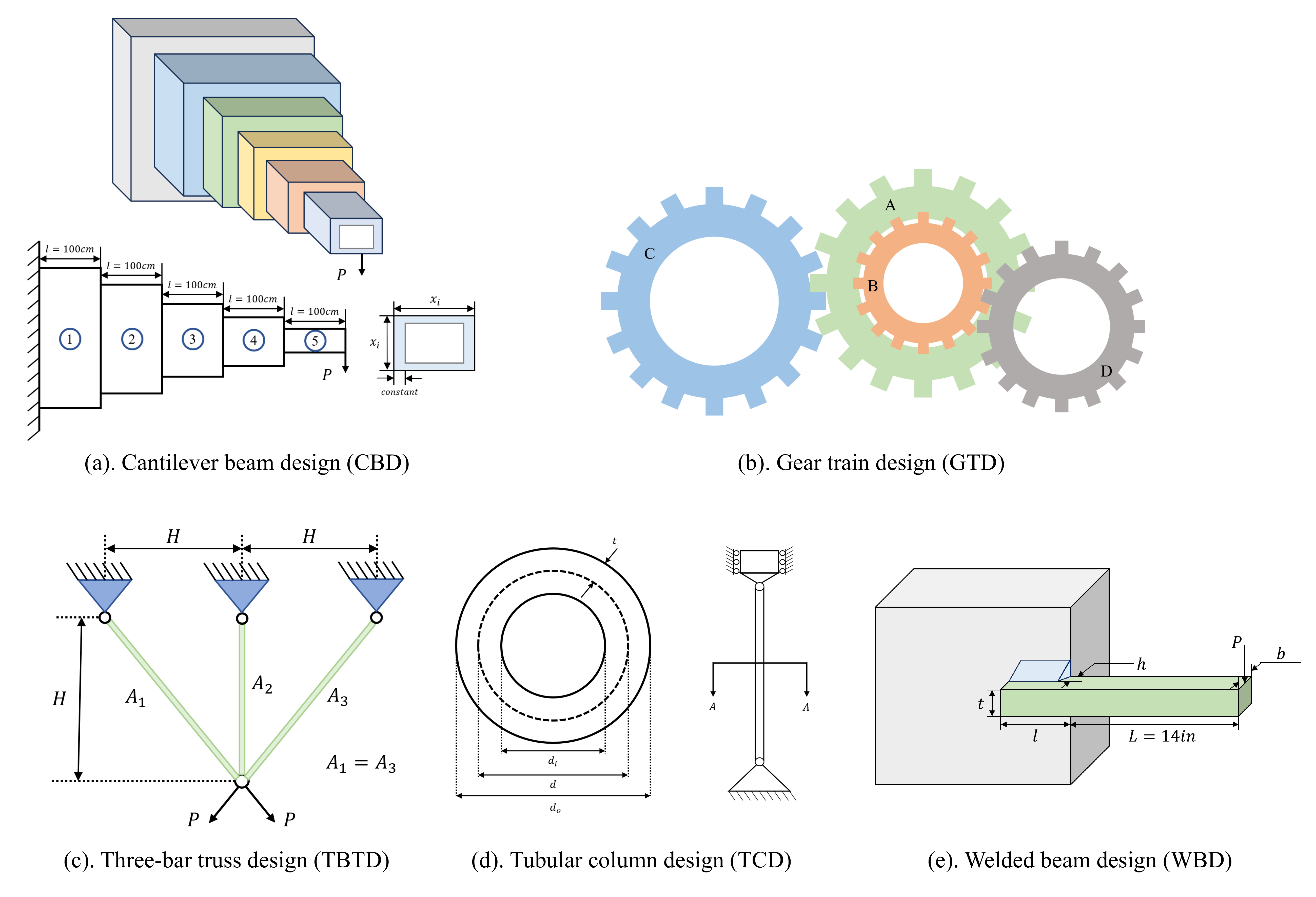}
    \caption{The demonstration of engineering simulation tasks.}
    \label{fig:2.2.1}
\end{figure*}

\textbf{Cantilever beam design problem (CBD)}: CBD aims to minimize the overall mass of the cantilever beam while ensuring it meets the specified bearing capacity requirements. Eq. (\ref{eq:2.2.1}) describes the mathematical model and Fig. \ref{fig:2.2.1}(a) presents a demonstration.
\begin{equation}
    \label{eq:2.2.1}
    \begin{aligned}
		& \textbf{min} \ f(X) = 0.0624 (x_1 + x_2 + x_3 + x_4 + x_5) \\
        & \textbf{s.t.} \ g(X) = \frac{61}{x^3_1} + \frac{37}{x^3_2} + \frac{19}{x^3_3} + \frac{7}{x^3_4} + \frac{1}{x^3_5} - 1 \leq 0 \\
        & \textbf{where} \ 0.01 \leq x_i \leq 100, \ i \in \{1, 2, 3, 4, 5\}
    \end{aligned} 
\end{equation}

\textbf{Corrugated bulkhead design (CBHD)}: CBHD aims to design a bulkhead that can efficiently resist certain forces or loads, in which the design variables are the width $x_1$, depth $x_2$, length $x_3$, and plate thickness $x_4$. The mathematical model of the CBHD is presented in Eq. (\ref{eq:2.2.2}).
\begin{equation}
    \label{eq:2.2.2}
    \begin{aligned}
		\textbf{min} \ f(X) = & \frac{5.885x_4(x_1 + x_3)}{x_1 + \sqrt{\vert x^2_3 - x^2_2\vert}} \\
        \textbf{s.t.} \ g_1(X) = & -x_4x_2(0.4x_1 + \frac{x_3}{6}) + \\
                                 & 8.94(x_1 + \sqrt{\vert x^2_3 - x^2_2\vert}) \leq 0 \\
        g_2(X) = & -x_4x^2_2(0.2x_1 + \frac{x_3}{12}) + \\ 
                 & 2.2(8.94(x_1 + \sqrt{\vert x^2_3 - x^2_2\vert}))^{4/3} \leq 0 \\
        g_3(X) = & -x_4 + 0.0156x_1 + 0.15 \leq 0 \\
        g_4(X) = & -x_4 + 0.0156x_3 + 0.15 \leq 0 \\
        g_5(X) = & -x_4 + 1.05 \leq 0 \\
        g_6(X) = & -x_3 + x_2 \leq 0 \\
        \textbf{where} \ 0 \leq & x_1, x_2, x_3 \leq 100 \\
        0 \leq & x_4 \leq 5 
    \end{aligned} 
\end{equation}

\textbf{Gear train design problem (GTD)}: GTD aims to minimize the gear ratio, defined as the ratio of the output shaft's angular velocity to the input shaft's angular velocity. The design variables include the number of teeth of gears $n_A=x_1$, $n_B=x_2$, $n_C=x_3$, and $n_D=x_4$, as expressed in Eq. (\ref{eq:2.2.3}) and demonstrated in Fig. \ref{fig:2.2.1}(b).
\begin{equation}
    \label{eq:2.2.3}
    \begin{aligned}
		& \textbf{min} \ f(X) = \left(\frac{1}{6.931} - \frac{x_3x_2}{x_1x_4}\right)^2 \\
        & \textbf{where} \ x_1, x_2, x_3, x_4 \in \{12, 13, 14, ..., 60\}
    \end{aligned} 
\end{equation}

\textbf{Three-bar truss design problem (TBTD)}: The objective of TBTD is to find the optimal configuration of a truss made up of three bars subject to the optimal cross-sectional areas $A_1=x_1$ and $A_2=x_2$. The mathematical model and demonstration are presented in Eq. (\ref{eq:2.2.4}) and Fig. \ref{fig:2.2.1}(c).
\begin{equation}
    \label{eq:2.2.4}
    \begin{aligned}
		\textbf{min} \ f(X) = & (2\sqrt{2}x_1 + x_2) \cdot l \\
        \textbf{s.t.} \ g_1(X) = & \frac{\sqrt{2}x_1 + x_2}{\sqrt{2}x^2_1 + 2x_1x_2}P-\sigma \leq 0 \\
        g_2(X) = & \frac{x_2}{\sqrt{2}x^2_1 + 2x_1x_2}P-\sigma \leq 0 \\
        g_3(X) = & \frac{1}{\sqrt{2}x_2 + x_1}P-\sigma \leq 0 \\
        l = & 100cm, P = 2kN/cm^3, \sigma = 2kN/cm^3 \\
        \textbf{where} \ 
        0 \leq & x_1, x_2 \leq 1 
    \end{aligned} 
\end{equation}

\textbf{Tubular column design problem (TCD)}: TCD through optimizing two decision variables: the mean diameter of the column $d=x_1$ and the thickness of tube $t=x_2$ to determine the optimum of a tubular column. Eq. (\ref{eq:2.2.5}) formulates the model and Fig. \ref{fig:2.2.1}(d) presents a demonstration.
\begin{equation}
    \label{eq:2.2.5}
    \begin{aligned}
		\textbf{min} \ f(X) = & 9.8x_1x_2 + 2x_1 \\
        \textbf{s.t.} \ g_1(X) = & \frac{P}{\pi x_1x_2\sigma_y}-1 \leq 0 \\
        g_2(X) = & \frac{8PL^2}{\pi^3Ex_1x_2(x^2_1+x^2_2)}-1 \leq 0 \\
        \textbf{where} \ 
        2 \leq & x_1 \leq 14 \\
        0.2 \leq & x_2 \leq 8
    \end{aligned} 
\end{equation}

\textbf{Welded beam design (WBD)}: The objective of WBD is to design a welded beam subjected to the weld thickness $h=x_1$, height $l=x_2$, length $t=x_3$, and bar thickness $b=x_4$, as formulated in Eq. (\ref{eq:2.2.6}) and visualized in Fig. \ref{fig:2.2.1}(e).
\begin{equation}
    \label{eq:2.2.6}
    \begin{aligned}
        \textbf{min} \ f(X) = & 1.10471x^2_1+0.04811x_3x_4(14+x_2) \\
        \textbf{s.t.} \ g_1(X) = & \tau(X) - \tau_{max} \leq 0 \\
        g_2(X) = & \sigma(X) - \sigma_{max} \leq 0 \\
        g_3(X) = & \theta(X) - \theta_{max} \leq 0 \\
        g_4(X) = & x_1-x_4 \leq 0 \\
        g_5(X) = & P-P_c(X) \leq 0 \\
        g_6(X) = & 0.125-x_1 \leq 0 \\
        g_6(X) = & 1.10471x^2_1+0.04811x_3x_4(14+x_2)-5 \leq 0 \\
        \textbf{where} \ 
        0.1 \leq & x_1, x_4 \leq 2 \\
        0.1 \leq & x_2, x_3 \leq 10 
    \end{aligned}
\end{equation}

\subsubsection{Competitor algorithms} \label{sec:4.1.2}
We compare CDE with state-of-the-art optimizers and DE variants. The specific algorithms and corresponding parameter settings are listed in Table \ref{tbl:4.1.2.1}.
\begin{table}[htbp]
	\scriptsize
	\centering
	\renewcommand\arraystretch{1.3}
	\caption{The compared optimizers and parameter settings}
	\label{tbl:4.1.2.1}
	\begin{tabular}{ccc}
		\toprule
		  Method & Parameters & Value \\
		\midrule
            \multirow{2}{*}{CDE} & $\mu_F$ and $\sigma_F$ & 0.5 and 0.3 \\
            ~ & $\mu_{Cr}$ and $\sigma_{Cr}$ & 0.5 and 0.3 \\
        \midrule
            \multirow{3}{*}{DE \cite{Storn:97}} & scale factor $F$ & 0.5 \\
            ~ & crossover rate $Cr$ & 0.8 \\
            ~ & mutation strategy & DE/rand/1/bin \\
        \midrule
            CMA-ES \cite{Hansen:01} & $\sigma$ & 1.3 \\
        \midrule
            \multirow{2}{*}{SaDE \cite{Qin:05}} & $\mu_F$ and $\sigma_F$ & 0.5 and 0.3 \\
            ~ & $\mu_{Cr}$ and $\sigma_{Cr}$ & 0.5 and 0.1 \\
        \midrule
            JADE \cite{Zhang:09} & $\mu_F$ and $\mu_{Cr}$ & 0.5 and 0.5 \\
        \midrule
            \multirow{2}{*}{L-SHADE \cite{Tanabe:14}} & population size $N$ & 18 $\times$ D \\
            ~ & $\mu_F$ and $\mu_{Cr}$ & 0.5 and 0.5 \\
        \midrule
            \multirow{3}{*}{L-SHADE-PWI \cite{Adam:18}} & population size $N$ & 18 $\times$ D \\
            ~ & $N_{min}$ & 4 \\
            ~ & $\mu_F$ and $\mu_{Cr}$ & 0.5 and 0.5 \\
        \midrule
            \multirow{2}{*}{GTDE \cite{Jia:23}} & $\mu_F$ and $\sigma_F$ & 0.7 and 0.5 \\
            ~ & $\mu_{Cr}$ and $\sigma_{Cr}$ & 0.5 and 0.3 \\
        \bottomrule
	\end{tabular}
\end{table}

Except for L-SHADE and L-SHADE-PWI, the population size for the rest of the algorithms is fixed at 100. The maximum fitness evaluation (FE) for CEC2017 benchmark functions and engineering simulation optimization tasks are set to 500 $\times$ D and 10,000, respectively. Each single algorithm is implemented with 30 trial runs to alleviate the effect of randomness. Additionally, the Holm multiple comparison test \cite{Holm:79} is employed to determine the significance between every pair of compared algorithms. The symbols $+$, $\approx$, and $-$ denote that the proposed CDE is significantly better, has no significant difference, and is significantly worse than the compared algorithm.

\subsection{Experimental results and statistical analyses} \label{sec:4.2}
\begin{figure*}[htbp]
    \centering
    \includegraphics[width=16cm]{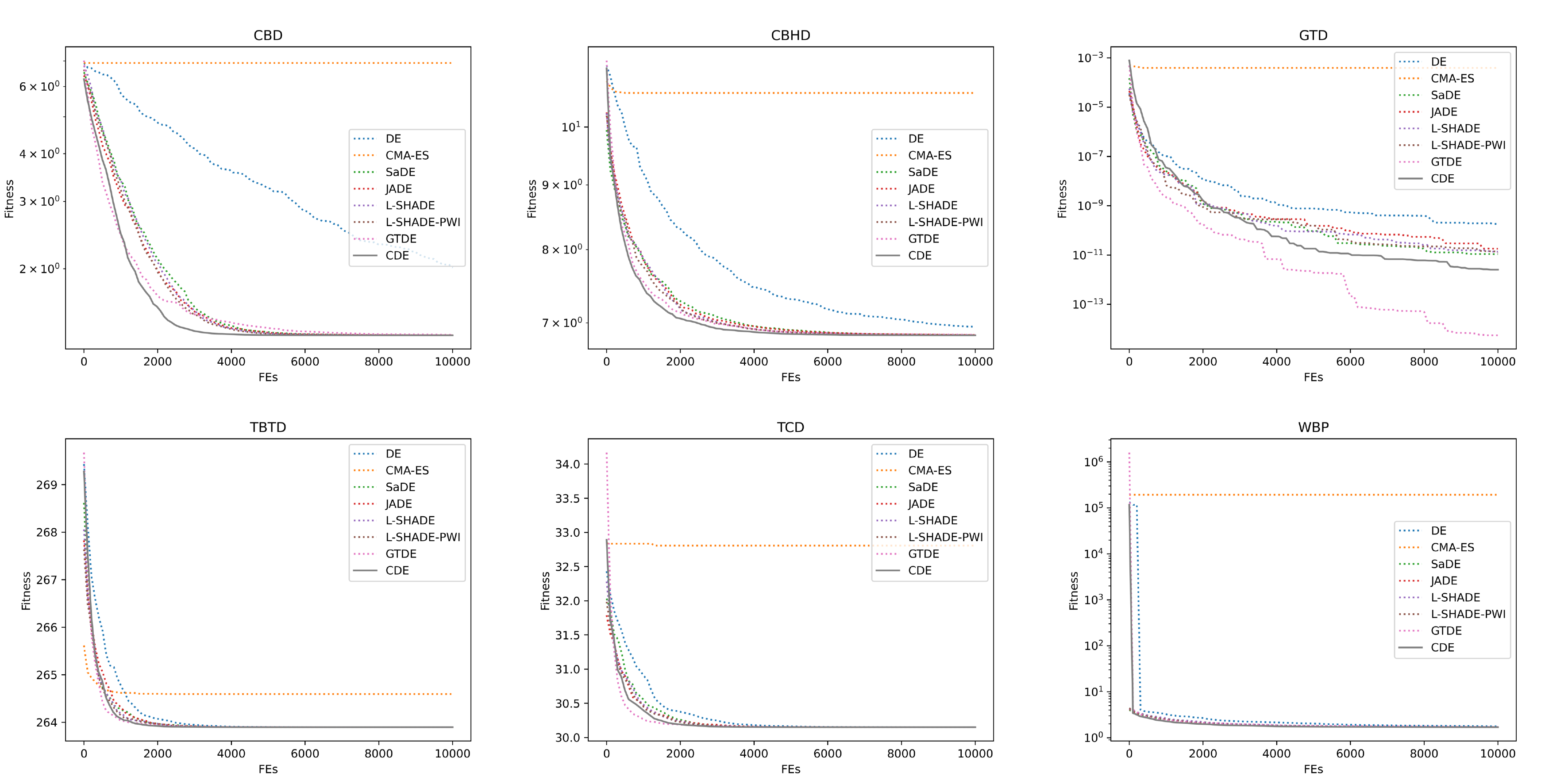}
    \caption{Convergence curves of eight algorithms on six engineering simulation optimization tasks.}
    \label{fig:4.2.1}
\end{figure*}

We summarize the experimental results and statistical analyses on the CEC2017 benchmark functions and engineering simulation optimization tasks in Tables \ref{tbl:4.2.1} and \ref{tbl:4.2.2}, respectively. The convergence curves on engineering tasks are demonstrated in Fig. \ref{fig:4.2.1}.
\begin{table*}[htbp]
	\centering
	\caption{Experimental results and statistical analyses on 30-D CEC2017. $f_1$: Unimodal function; $f_3-f_9$: Simple multimodal functions; $f_{10}-f_{19}$: Hybrid functions; $f_{20}-f_{30}$: Composite functions.}
	\renewcommand\arraystretch{1.2}
	\label{tbl:4.2.1}
	\resizebox{0.8\linewidth}{!}{
	\begin{tabular}{cccccccccc}
		\toprule
		\multicolumn{2}{c}{Func.} & DE & CMA-ES & SaDE & JADE & L-SHADE & L-SHADE-PWI & GTDE & CDE \\
		\midrule
        \multirow{2}{*}{$f_1$} & mean & 5.103e+10 $+$ & 3.727e+10 $+$ & 6.658e+06 $+$ & 2.557e+05 $+$ & 2.247e+05 $+$ & 2.005e+05 $+$ & 4.846e+05 $+$ & \textbf{3.935e+03} \\
        ~ & std & 4.548e+09 & 4.131e+09 & 3.586e+06 & 1.062e+05 & 8.170e+04 & 9.129e+04 & 7.482e+05 & 2.858e+03 \\
        \midrule
        \multirow{2}{*}{$f_3$} & mean & 2.570e+05 $+$ & 8.737e+04 $+$ & 1.256e+05 $+$ & 1.097e+05 $+$ & 7.925e+04 $\approx$ & 9.514e+04 $+$ & \textbf{5.365e+04} $-$ & 6.566e+04 \\
        ~ & std & 3.839e+04 & 1.087e+04 & 1.324e+04 & 1.553e+04 & 2.571e+04 & 2.334e+04 & 1.578e+04 & 1.017e+04 \\
        \multirow{2}{*}{$f_4$} & mean & 5.598e+03 $+$ & 1.040e+04 $+$ & 5.155e+02 $+$ & 4.971e+02 $\approx$ & 5.119e+02 $+$ & 5.069e+02 $\approx$ & 5.229e+02 $+$ & \textbf{4.947e+02} \\
        ~ & std & 1.001e+03 & 1.669e+03 & 1.121e+01 & 9.735e+00 & 1.317e+01 & 1.516e+01 & 5.260e+01 & 2.754e+01 \\
        \multirow{2}{*}{$f_5$} & mean & 9.157e+02 $+$ & 8.647e+02 $+$ & 7.067e+02 $+$ & 6.830e+02 $\approx$ & 6.812e+02 $\approx$ & 6.961e+02 $+$ & \textbf{6.188e+02} $-$ & 6.793e+02 \\
        ~ & std & 1.467e+01 & 2.180e+01 & 8.716e+00 & 1.230e+01 & 8.704e+00 & 1.070e+01 & 2.373e+01 & 2.163e+01 \\
        \multirow{2}{*}{$f_6$} & mean & 6.854e+02 $+$ & 6.841e+02 $+$ & 6.039e+02 $+$ & 6.030e+02 $+$ & 6.017e+02 $+$ & 6.016e+02 $+$ & 6.094e+02 $+$ & \textbf{6.000e+02} \\
        ~ & std & 4.509e+00 & 6.360e+00 & 6.699e-01 & 6.214e-01 & 2.573e-01 & 1.715e-01 & 6.514e+00 & 6.747e-02 \\
        \multirow{2}{*}{$f_7$} & mean & 2.885e+03 $+$ & 1.316e+03 $+$ & 9.519e+02 $+$ & 9.257e+02 $+$ & 9.164e+02 $\approx$ & 9.245e+02 $\approx$ & \textbf{9.008e+02} $\approx$ & 9.177e+02 \\
        ~ & std & 1.479e+02 & 5.799e+01 & 1.039e+01 & 1.219e+01 & 1.091e+01 & 1.012e+01 & 4.178e+01 & 1.398e+01 \\
        \multirow{2}{*}{$f_8$} & mean & 1.225e+03 $+$ & 1.108e+03 $+$ & 1.010e+03 $+$ & 9.816e+02 $\approx$ & 9.784e+02 $-$ & 9.940e+02 $\approx$ & \textbf{9.204e+02} $-$ & 9.891e+02 \\
        ~ & std & 2.398e+01 & 2.175e+01 & 1.576e+01 & 1.171e+01 & 9.673e+00 & 1.037e+01 & 3.005e+01 & 1.463e+01 \\
        \multirow{2}{*}{$f_9$} & mean & 2.034e+04 $+$ & 9.916e+03 $+$ & 1.080e+03 $+$ & 9.334e+02 $+$ & 9.227e+02 $+$ & 9.204e+02 $+$ & 2.833e+03 $+$ & \textbf{9.121e+02} \\
        ~ & std & 3.367e+03 & 1.739e+03 & 3.154e+01 & 1.779e+01 & 6.416e+00 & 7.894e+00 & 1.252e+03 & 2.395e+01 \\
        \midrule
        \multirow{2}{*}{$f_{10}$} & mean & 8.535e+03 $+$ & 8.965e+03 $+$ & 8.526e+03 $+$ & 7.857e+03 $-$ & 7.788e+03 $-$ & 8.264e+03 $\approx$ & \textbf{6.230e+03} $-$ & 8.288e+03 \\
        ~ & std & 1.251e+02 & 3.064e+02 & 2.501e+02 & 5.205e+02 & 4.083e+02 & 4.103e+02 & 1.218e+03 & 3.137e+02 \\
        \multirow{2}{*}{$f_{11}$} & mean & 5.459e+03 $+$ & 7.084e+03 $+$ & 1.580e+03 $+$ & 1.418e+03 $+$ & 1.330e+03 $+$ & 1.331e+03 $+$ & 1.336e+03 $+$ & \textbf{1.211e+03} \\
        ~ & std & 9.216e+02 & 1.281e+03 & 7.087e+01 & 1.369e+02 & 2.471e+01 & 1.982e+01 & 7.101e+01 & 3.607e+01 \\
        \multirow{2}{*}{$f_{12}$} & mean & 2.890e+09 $+$ & 9.017e+09 $+$ & 8.847e+06 $+$ & 2.351e+06 $+$ & 2.646e+06 $+$ & 2.455e+06 $+$ & 1.394e+06 $+$ & \textbf{3.315e+05} \\
        ~ & std & 7.294e+08 & 1.469e+09 & 3.007e+06 & 1.061e+06 & 9.611e+05 & 1.091e+06 & 1.272e+06 & 2.678e+05 \\
        \multirow{2}{*}{$f_{13}$} & mean & 3.517e+08 $+$ & 8.504e+09 $+$ & 1.856e+06 $+$ & 7.919e+05 $+$ & 8.107e+05 $+$ & 6.539e+05 $+$ & 1.891e+05 $+$ & \textbf{1.791e+04} \\
        ~ & std & 1.550e+08 & 2.622e+09 & 1.525e+06 & 4.738e+05 & 3.963e+05 & 3.435e+05 & 8.023e+05 & 1.141e+04 \\
        \multirow{2}{*}{$f_{14}$} & mean & 1.545e+05 $+$ & 3.709e+06 $+$ & 4.919e+04 $+$ & 5.594e+04 $+$ & 9.363e+03 $+$ & 1.253e+04 $+$ & \textbf{3.568e+03} $-$ & 5.788e+03 \\
        ~ & std & 8.448e+04 & 3.375e+06 & 2.024e+04 & 1.000e+05 & 3.375e+03 & 6.404e+03 & 3.773e+03 & 4.171e+03 \\
        \multirow{2}{*}{$f_{15}$} & mean & 1.099e+07 $+$ & 1.057e+08 $+$ & 2.662e+05 $+$ & 1.879e+05 $+$ & 7.902e+04 $+$ & 8.338e+04 $+$ & 1.527e+04 $+$ & \textbf{7.296e+03} \\
        ~ & std & 5.523e+06 & 7.358e+07 & 2.447e+05 & 1.815e+05 & 3.691e+04 & 2.115e+04 & 1.292e+04 & 4.826e+03 \\
        \multirow{2}{*}{$f_{16}$} & mean & 4.100e+03 $+$ & 6.112e+03 $+$ & 3.283e+03 $\approx$ & 3.146e+03 $\approx$ & 3.137e+03 $\approx$ & 3.202e+03 $\approx$ & \textbf{2.717e+03} $-$ & 3.121e+03 \\
        ~ & std & 2.520e+02 & 7.003e+02 & 2.261e+02 & 1.656e+02 & 2.096e+02 & 1.460e+02 & 2.888e+02 & 1.890e+02 \\
        \multirow{2}{*}{$f_{17}$} & mean & 2.902e+03 $+$ & 3.591e+03 $+$ & 2.302e+03 $+$ & 2.278e+03 $+$ & 2.201e+03 $+$ & 2.202e+03 $+$ & 2.241e+03 $+$ & \textbf{2.071e+03} \\
        ~ & std & 1.397e+02 & 4.381e+02 & 1.375e+02 & 1.029e+02 & 9.999e+01 & 8.212e+01 & 1.897e+02 & 1.211e+02 \\
        \multirow{2}{*}{$f_{18}$} & mean & 8.635e+06 $+$ & 2.787e+07 $+$ & 2.636e+06 $+$ & 3.552e+05 $-$ & 8.008e+05 $\approx$ & 9.985e+05 $+$ & \textbf{2.694e+05} $-$ & 6.445e+05 \\
        ~ & std & 2.858e+06 & 1.867e+07 & 1.543e+06 & 7.014e+05 & 3.289e+05 & 3.210e+05 & 3.029e+05 & 2.997e+05 \\
        \multirow{2}{*}{$f_{19}$} & mean & 5.534e+07 $+$ & 1.479e+08 $+$ & 3.885e+05 $+$ & 2.783e+05 $+$ & 1.231e+05 $+$ & 1.098e+05 $+$ & 2.082e+04 $\approx$ & \textbf{1.071e+04} \\
        ~ & std & 1.531e+07 & 6.526e+07 & 1.993e+05 & 3.473e+05 & 5.324e+04 & 4.258e+04 & 2.342e+04 & 8.252e+03 \\
        \midrule
        \multirow{2}{*}{$f_{20}$} & mean & 2.895e+03 $+$ & 2.980e+03 $+$ & 2.753e+03 $+$ & 2.645e+03 $+$ & 2.685e+03 $+$ & 2.679e+03 $+$ & \textbf{2.422e+03} $-$ & 2.574e+03 \\
        ~ & std & 1.522e+02 & 1.782e+02 & 1.530e+02 & 1.059e+02 & 9.051e+01 & 1.390e+02 & 1.717e+02 & 1.038e+02 \\
        \multirow{2}{*}{$f_{21}$} & mean & 2.703e+03 $+$ & 2.702e+03 $+$ & 2.497e+03 $+$ & 2.482e+03 $+$ & 2.477e+03 $\approx$ & 2.480e+03 $\approx$ & \textbf{2.421e+03} $-$ & 2.473e+03 \\
        ~ & std & 1.745e+01 & 3.178e+01 & 1.541e+01 & 1.360e+01 & 7.071e+00 & 2.119e+01 & 2.781e+01 & 1.477e+01 \\
        \multirow{2}{*}{$f_{22}$} & mean & 1.007e+04 $+$ & 9.432e+03 $+$ & 4.272e+03 $+$ & 3.152e+03 $+$ & 2.380e+03 $+$ & 2.332e+03 $+$ & 6.922e+03 $+$ & \textbf{2.302e+03} \\
        ~ & std & 2.811e+02 & 6.399e+02 & 2.839e+03 & 2.109e+03 & 1.794e+02 & 4.572e+01 & 2.150e+03 & 4.604e+00 \\
        \multirow{2}{*}{$f_{23}$} & mean & 3.048e+03 $+$ & 3.864e+03 $+$ & 2.850e+03 $+$ & 2.845e+03 $+$ & 2.835e+03 $+$ & 2.836e+03 $+$ & \textbf{2.791e+03} $\approx$ & 2.800e+03 \\
        ~ & std & 2.042e+01 & 1.486e+02 & 1.184e+01 & 1.235e+01 & 1.104e+01 & 1.078e+01 & 2.802e+01 & 3.928e+01 \\
        \multirow{2}{*}{$f_{24}$} & mean & 3.168e+03 $+$ & 4.123e+03 $+$ & 3.021e+03 $+$ & 3.014e+03 $+$ & 3.003e+03 $+$ & 3.002e+03 $+$ & \textbf{2.965e+03} $-$ & 2.981e+03 \\
        ~ & std & 1.452e+01 & 1.231e+02 & 1.162e+01 & 1.262e+01 & 9.793e+00 & 1.587e+01 & 3.329e+01 & 3.151e+01 \\
        \multirow{2}{*}{$f_{25}$} & mean & 7.916e+03 $+$ & 4.393e+03 $+$ & 2.896e+03 $\approx$ & \textbf{2.888e+03} $\approx$ & 2.889e+03 $\approx$ & 2.890e+03 $\approx$ & 2.908e+03 $+$ & 2.896e+03 \\
        ~ & std & 7.414e+02 & 2.358e+02 & 5.233e+00 & 1.622e+00 & 1.329e+00 & 3.488e+00 & 2.653e+01 & 1.487e+01 \\
        \multirow{2}{*}{$f_{26}$} & mean & 8.009e+03 $+$ & 1.022e+04 $+$ & 5.733e+03 $+$ & 5.441e+03 $+$ & 5.381e+03 $+$ & 5.432e+03 $+$ & 5.082e+03 $+$ & \textbf{4.319e+03} \\
        ~ & std & 1.622e+02 & 4.395e+02 & 1.563e+02 & 1.552e+02 & 1.210e+02 & 1.653e+02 & 6.734e+02 & 9.410e+02 \\
        \multirow{2}{*}{$f_{27}$} & mean & 3.319e+03 $+$ & 5.137e+03 $+$ & 3.237e+03 $\approx$ & \textbf{3.227e+03} $-$ & 3.229e+03 $\approx$ & 3.231e+03 $\approx$ & 3.245e+03 $\approx$ & \textbf{3.234e+03} \\
        ~ & std & 1.968e+01 & 3.727e+02 & 6.785e+00 & 4.321e+00 & 3.571e+00 & 7.636e+00 & 2.469e+01 & 1.278e+01 \\
        \multirow{2}{*}{$f_{28}$} & mean & 5.625e+03 $+$ & 6.320e+03 $+$ & 3.273e+03 $+$ & 3.243e+03 $\approx$ & 3.253e+03 $\approx$ & 3.244e+03 $\approx$ & 3.374e+03 $+$ & \textbf{3.235e+03} \\
        ~ & std & 7.238e+02 & 4.928e+02 & 1.409e+01 & 2.169e+01 & 1.408e+01 & 1.688e+01 & 2.394e+02 & 2.565e+01 \\
        \multirow{2}{*}{$f_{29}$} & mean & 4.892e+03 $+$ & 7.512e+03 $+$ & 4.229e+03 $+$ & 4.071e+03 $+$ & 4.070e+03 $+$ & 4.122e+03 $+$ & 4.009e+03 $+$ & \textbf{3.846e+03} \\
        ~ & std & 2.164e+02 & 6.220e+02 & 2.283e+02 & 1.605e+02 & 1.524e+02 & 1.161e+02 & 2.031e+02 & 1.855e+02 \\
        \multirow{2}{*}{$f_{30}$} & mean & 3.759e+07 $+$ & 1.003e+09 $+$ & 1.192e+06 $+$ & 3.945e+05 $+$ & 4.913e+05 $+$ & 5.271e+05 $+$ & 1.283e+05 $\approx$ & \textbf{5.083e+04} \\
        ~ & std & 1.225e+07 & 4.249e+08 & 7.053e+05 & 2.741e+05 & 1.892e+05 & 1.731e+05 & 2.266e+05 & 3.913e+04 \\
        \midrule
        \multicolumn{2}{c}{$+$/$\approx$/$-$:} & 29/0/0 & 29/0/0 & 26/3/0 & 20/6/3 & 18/9/2 & 20/9/0 & 14/5/10 & - \\
        \midrule
        \multicolumn{2}{c}{Avg. rank:} & 7.3 & 7.5 & 5.7 & 3.8 & 3.2 & 3.7 & 2.9 & \textbf{1.9} \\
        \bottomrule
	\end{tabular}
   }
\end{table*}

\begin{table*}[htbp]
	\centering
	\caption{Experimental results and statistical analyses on engineering optimization problems.}
	\renewcommand\arraystretch{1.2}
	\label{tbl:4.2.2}
	\resizebox{0.85\linewidth}{!}{
	\begin{tabular}{cccccccccc}
		\toprule
		\multicolumn{2}{c}{Func.} & DE & CMA-ES & SaDE & JADE & L-SHADE & L-SHADE-PWI & GTDE & CDE \\
		\midrule
        \multirow{2}{*}{CBD} & mean & 2.015e+00 $+$ & 6.910e+00 $+$ & 1.341e+00 $+$ & 1.341e+00 $+$ & 1.340e+00 $+$ & 1.340e+00 $+$ & 1.344e+00 $+$ & \textbf{1.340e+00} \\
        ~ & std & 2.139e-01 & 1.248e+00 & 3.301e-04 & 3.767e-04 & 1.935e-04 & 1.705e-04 & 2.207e-03 & 3.705e-05 \\
		\midrule
        \multirow{2}{*}{CBHD} & mean & 6.949e+00 $+$ & 1.064e+01 $+$ & 6.845e+00 $+$ & 6.847e+00 $+$ & 6.844e+00 $+$ & 6.844e+00 $+$ & 6.850e+00 $+$ & \textbf{6.843e+00} \\
        ~ & std & 3.671e-02 & 1.251e+00 & 1.237e-03 & 1.529e-03 & 3.915e-04 & 5.784e-04 & 3.810e-03 & 3.375e-04 \\
		\midrule
        \multirow{2}{*}{GTD} & mean & 1.838e-10 $+$ & 3.924e-04 $+$ & 1.087e-11 $+$ & 1.805e-11 $+$ & 1.373e-11 $+$ & 1.200e-11 $+$ & \textbf{5.568e-15} $-$ & 2.560e-12 \\
        ~ & std & 2.820e-10 & 7.737e-04 & 1.283e-11 & 3.519e-11 & 2.412e-11 & 2.425e-11 & 1.670e-14 & 5.844e-12 \\
		\midrule
        \multirow{2}{*}{TBTD} & mean & 2.639e+02 $+$ & 2.646e+02 $+$ & 2.639e+02 $+$ & 2.639e+02 $+$ & 2.639e+02 $+$ & 2.639e+02 $+$ & 2.639e+02 $+$ & \textbf{2.639e+02} \\
        ~ & std & 7.558e-06 & 6.359e-01 & 7.622e-07 & 1.574e-06 & 3.310e-07 & 4.045e-07 & 1.063e-04 & 5.984e-08 \\
		\midrule
        \multirow{2}{*}{TCD} & mean & 3.015e+01 $+$ & 3.281e+01 $+$ & 3.015e+01 $+$ & 3.015e+01 $+$ & 3.015e+01 $+$ & 3.015e+01 $+$ & 3.015e+01 $+$ & \textbf{3.015e+01} \\
        ~ & std & 8.503e-05 & 1.439e+00 & 2.198e-06 & 1.019e-05 & 2.300e-06 & 2.962e-06 & 1.013e-04 & 5.172e-07 \\
		\midrule
        \multirow{2}{*}{WBP} & mean & 1.761e+00 $+$ & 1.932e+05 $+$ & 1.696e+00 $+$ & 1.692e+00 $+$ & 1.690e+00 $+$ & 1.689e+00 $+$ & 1.712e+00 $+$ & \textbf{1.687e+00} \\
        ~ & std & 2.811e-02 & 1.040e+06 & 9.128e-03 & 3.993e-03 & 2.950e-03 & 2.224e-03 & 4.087e-02 & 5.447e-03 \\
        \midrule
        \multicolumn{2}{c}{$+$/$\approx$/$-$:} & 6/0/0 & 6/0/0 & 6/0/0 & 6/0/0 & 6/0/0 & 6/0/0 & 5/0/1 & - \\
        \midrule
        \multicolumn{2}{c}{Avg. rank:} & 6.7 & 8.0 & 4.0 & 5.0 & 2.8 & 2.8 & 5.5 & 1.2 \\
        \bottomrule
	\end{tabular}
   }
\end{table*}

\section{Discussion}  \label{sec:5}

\subsection{Performance analysis on CEC2017}  \label{sec:5.1}
Since the CEC2017 benchmark suite contains test functions with various characteristics such as unimodal, multimodal, hybrid, and composite, thus the optimization in these test functions can fully reflect the performance of optimizers and support us in investigating the features of involved algorithms thoroughly. 

Initially, $f_1$ is unimodal functions, and the optimization in these functions allows the performance evaluation in the aspect of the exploitative capacity. The superiority of CDE is apparent in CEC2017 $f_1$ compared with state-of-the-art optimizers. Therefore, we conclude that CDE has a remarkable exploitation ability and robust performance across various problem domains. 

Subsequently, $f_3$ to $f_9$ are multimodal functions. These functions contain more than one local optima and evaluate the performance of optimizers in escaping from local optima and global convergence. Through the experimental results and statistical analyses summarized in Table \ref{tbl:4.2.1}, the competitiveness of our proposed CDE is observable. As the state-of-the-art DE variant, GTDE outperforms CDE in some instances such as $f_3$, $f_5$, and $f_8$. However, the excellent performance of CDE cannot be neglected. Overall, CDE best performs in $f_4$, $f_6$, and $f_9$, and the capacities in escaping from local optima and global convergence are experimentally verified through the results.

Finally, the rest of the functions are hybrid and composite. These functions have complex fitness landscapes and multiple optima, which challenges the abilities of optimizers in balancing exploitation and exploration, avoiding premature convergence, and achieving global optimization. Upon review of the result summary, it becomes evident that CDE consistently demonstrates superior performance across many test functions within this category, thereby highlighting its efficacy in complex optimization environments.

\subsection{Performance analysis on engineering tasks}  \label{sec:5.2}
The engineering simulation optimization tasks serve as real-world challenges to evaluate the performance of optimizers in complex optimization scenarios. This study introduces CDE as a novel approach to deal with engineering optimization tasks. Remarkably, our proposed CDE outperforms all other methods across all instances except for GTD when compared with GTDE, showcasing its superior performance in this domain. 

In summary, our proposed CDE is a satisfactory variant of DE in both benchmark and engineering optimization. We owe this success to the integration of the competitive mechanism and the intelligent hyper-parameter adaptation inherited from the previous research \cite{Qin:05, Jia:23}. These elements collectively empower CDE  with outstanding efficiency and effectiveness.

\section{Conclusion}  \label{sec:6}
This paper proposes a novel competitive DE (CDE) to solve numerical optimization problems. We introduce a competitive mechanism to DE and propose a novel DE/winner-to-best/1 mutation strategy. Moreover, CDE inherits the hyper-parameter adaptation schemes recommended in \cite{Qin:05, Jia:23}. To assess the performance of CDE, we conduct comprehensive numerical experiments on CEC2017 benchmark functions and engineering simulation optimization problems. The experimental results and statistical analyses confirm the competitiveness of our proposed CDE compared to state-of-the-art EAs and advanced variants of DE, including CMA-ES, JADE, L-SHADE, L-SHADE-PWI, and GTDE. 

In conclusion, our proposed CDE exhibits significant potential as a powerful optimizer in real-world scenarios. In future research, we plan to further develop CDE and leverage its capabilities to address complex tasks across various application domains.

\section{Acknowledgement} \label{sec:7}
This work was supported by JSPS KAKENHI Grant Number 21A402 and JST SPRING Grant Number JPMJSP2119.

\bibliographystyle{IEEEtran}
\bibliography{paper}
\end{document}